\documentclass[sn-mathphys,Numbered]{sn-jnl}

\usepackage{natbib}
\usepackage{caption}

\usepackage[utf8]{inputenc} 
\usepackage{booktabs}       
\usepackage{amsmath,amssymb,amsfonts}
\usepackage{nicefrac}       
\usepackage[letterspace=-50]{microtype}      
\usepackage{mathtools}
\usepackage{enumitem}
\usepackage{amsthm}
\usepackage{subcaption}
\usepackage{url}   
\usepackage{graphicx}
\usepackage{hyperref}
\usepackage{orcidlink}

\usepackage{multirow}%
\usepackage{mathrsfs}%
\usepackage[title]{appendix}%
\usepackage{xcolor}%
\usepackage{textcomp}%
\usepackage{manyfoot}%
\usepackage{algorithm}%
\usepackage{algorithmicx}%
\usepackage{algpseudocode}%
\usepackage{listings}%

\makeatletter
\algrenewcommand\ALG@beginalgorithmic{\tiny\lsstyle}
\makeatother

\newcommand{\softmax}{\mathop{\mathrm{softmax}}}
\newcommand{\expect}{\mathop{\mathbb{E}}}

\algnewcommand{\IfThenElse}[3]{
  \State \algorithmicif\ #1\ \algorithmicthen\ #2\ \algorithmicelse\ #3}

\algnewcommand{\IfThen}[2]{
  \State \algorithmicif\ #1\ \algorithmicthen\ #2}

\theoremstyle{thmstyleone}%
\theoremstyle{thmstyletwo}%

\theoremstyle{thmstylethree}%

\begin{document}
\title{\centering Symbolic~Relational Deep~Reinforcement~Learning based~on~Graph~Neural~Networks and~Autoregressive~Policy~Decomposition}

\author*{\fnm{Jaromír} \sur{Janisch}\,\orcidlink{0000-0002-4165-6503}}\email{jaromir.janisch@fel.cvut.cz}
\author{\fnm{Tomáš} \sur{Pevný}\,\orcidlink{0000-0002-5768-9713}}\email{tomas.pevny@fel.cvut.cz}
\author{\fnm{Viliam} \sur{Lisý}\,\orcidlink{0000-0002-1647-1507}}\email{viliam.lisy@fel.cvut.cz}

\affil{\orgdiv{Artificial Intelligence Center, Department of Computer Science, \\ Faculty of Electrical Engineering}, \orgname{Czech Technical University in Prague}, \orgaddress{\street{Karlovo namesti 293/13}, \city{Prague}, \postcode{120 00}, \country{Czech Republic}}}

\abstract{We focus on reinforcement learning (RL) in relational problems that are naturally defined in terms of objects, their relations, and object-centric actions. These problems are characterized by variable state and action spaces, and finding a fixed-length representation, required by most existing RL methods, is difficult, if not impossible. We present a deep RL framework based on graph neural networks and auto-regressive policy decomposition that naturally works with these problems and is completely domain-independent. We demonstrate the framework's broad applicability in three distinct domains and show impressive zero-shot generalization over different problem sizes.}


\maketitle

\section{Introduction}

Current Deep Reinforcement Learning (Deep RL) focuses mostly on visual-control domains \citep[e.g.,][]{leibo2018psychlab,mnih2015human,jaderberg2019human} with fixed state and action space dimensions. However, for many real-world tasks, it is natural and easier to represent a domain in terms of \emph{objects}, their \emph{relations}, and \emph{actions} that directly manipulate them, rather than tensors. Such tasks naturally occur all around us. 

Let us imagine an internet bot that browses the web, uses different APIs and gathers pieces of information. It is natural to represent the data it is working with in the form of an ontology, which is a graph of entities where the connections represent relations. On the other hand, the same data would be difficult to translate into any fixed form, e.g., into a visual representation or general tensors. Similarly, the action space is also better represented using the objects (e.g., to make an API call to a newly discovered service), rather than a pre-described set of actions. 

\begin{figure}[t]
  \centering
  \begin{tabular}{ccc}
    \includegraphics[width=0.3\linewidth]{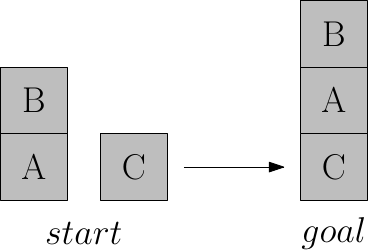} &
    \includegraphics[width=0.33\linewidth]{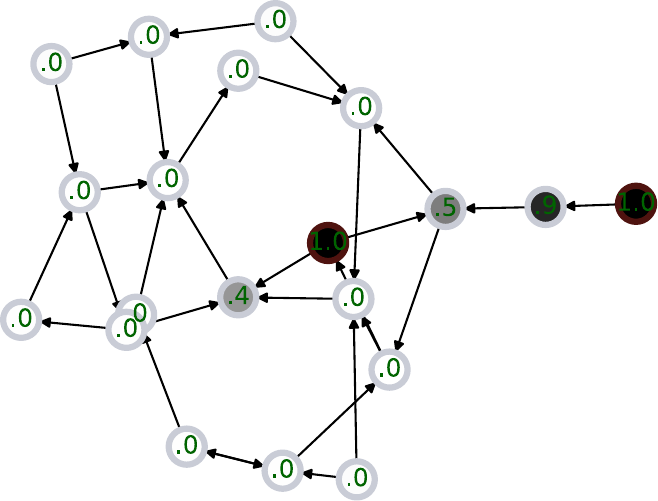} &
    \includegraphics[width=0.22\linewidth]{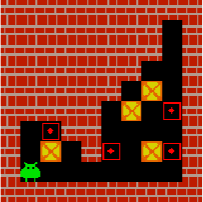} \\

    \emph{move}$(x, y)$ & 
    \emph{reset}$(x)$  or \emph{reset}${(\mathcal X)}$ &
    \emph{push-left}$(x)$, and  \\ 
    & & \emph{-right}, \emph{-down}, \emph{-up} \\

    (a) BlockWorld &
    (b) SysAdmin &
    (c) Sokoban 
  \end{tabular}

  \caption{Visualizations and actions of three domains where we demonstrate our method. In the BlockWorld game, the task is to reconfigure blocks into a goal position. In SysAdmin, computers in a network where nodes need to be selectively restarted to keep them running. In a variation of this environment, multiple nodes can be selected at once. Sokoban is a classic planning domain, where we include a twist -- actions are applied directly on the boxes. In all three domains, \textbf{the key challenge is to zero-shot generalize to different problem sizes}.}
  \label{fig:domains}
\end{figure}

Relational Reinforcement Learning (RRL) \citep{dvzeroski2001relational} studies tasks described in a relational language, in which states are described with predicates, such as \emph{on}$(x, y)$, and actions manipulate the objects, e.g., \emph{move}$(x, y)$. 
As an illustration, look at the BlockWorld problem \citep{slaney2001blocks} in Figure~\ref{fig:domains}a. Initially, several labeled blocks are stacked on top of each other in an arbitrary configuration. The task is to reconfigure them into a goal position, using a \emph{move}$(x, y)$ action that picks a block $x$ and puts it on top of $y$, which could also be the ground. 

In this work, we present a generic framework to solve a large class of relational problems using Deep RL. We call it SR-DRL (Symbolic Relational Deep RL), and it is designed to work with an enriched symbolic input (i.e., objects, their relations and their features) in the form of a graph and multi-parameter actions that target the objects. Being a Deep RL-based framework, it does not require the knowledge of transition dynamics and hence, it can be applied in domains where these are not available -- unlike planning, where a precise domain description is needed.

We use a Graph Neural Network (GNN) to process the input state, and decompose the policy into a sequence of parameter selections. This decomposition allows us to select an action in a linear time, wrt. the number of action parameters, where each of these parameters represents a specific node in the graph. A similar decomposition was studied by \citet{vinyals2017starcraft} who assumed that the parameters can be selected independently. However, not all probability distributions can be represented in this manner. Unlike \citep{vinyals2017starcraft}, we respect the parameters' conditional dependency. For example, in action \emph{move(x, y)}, the choice of \emph{y} should be dependent on the chosen \emph{x}. 

Additionally, we allow \emph{set} parameters, where any combination of objects can be independently selected. For example, a hypothetical action \emph{select}$(\mathcal X)$ with a set parameter $\mathcal X$ could select any set of input objects, with any cardinality. We train the policy using the A2C algorithm \citep{mnih2016asynchronous}, but any algorithm from the policy gradient family \citep{li2018deep} could be used (e.g., PPO \citep{schulman2017proximal}). The important property of the SR-DRL framework is that the \textbf{trained models are not constrained to the specific problem sizes, but can be used with a different number of objects and actions} (e.g., in the Sokoban environment, a model can be trained on 10×10 instances, but deployed to larger problems).

We demonstrate different aspects of our framework in three distinct domains. \textbf{a)} BlockWorld is a well-known planning domain with NP-hard complexity for optimal planning. We use its formulation with a two-parameter \emph{move}$(x, y)$ action to demonstrate how to use multiple parameters with a conditional dependency. Moreover, we show that an agent trained with only five blocks can be seamlessly deployed in a problem with 20 blocks and solve it with a 78\% success rate. \textbf{b)} Sokoban is a game requiring extensive planning and we use it to show how to manipulate the game's objects on the macro level, while a low-level planner translates the macro actions to a number of environment steps. The agent trained in 10×10 problems with four boxes solves 89\% of random 15×15 problems with five boxes. \textbf{c)} SysAdmin is a graph-based planning domain. In its variation \emph{SysAdmin-M}, we demonstrate our framework's capability to independently select a set of nodes at once. The agent trained with ten nodes generalizes almost perfectly to 160 nodes and performs comparably to PROST planner \citep{keller2012prost}, using only a fraction of decision time.


The following summarizes the main contributions of this paper:
\begin{itemize}
  \item We reformulate the relational reinforcement learning to operate on a graph, where nodes represent objects, edges their relations and the actions sequentially select multiple nodes to become action parameters. 
  \item We design a method based on a GNN, policy decomposition and deep reinforcement learning to learn a policy that maximizes designed problem reward. The action selection operates in linear time and space, wrt. the number of parameters. Additionally, the trained models are applicable to problem instances with a different number of objects.
  \item We provide an open-source implementation\footnote{The code is at \href{https://github.com/jaromiru/sr-drl}{\texttt{github.com/jaromiru/sr-drl}}.} and demonstrate the method in three distinct domains, showing impressive generalization capabilities. 
\end{itemize}
Additionally, the following summarizes the minor contributions:
\begin{itemize}
  \item In Sokoban, we demonstrate how to combine our method with a low-level controller that translates macro actions issued by the agent into elementary actions for the environment.
  \item We show how to modify the auto-regressive policy introduced in \citet{vinyals2017starcraft} for conditionally dependent parameters.
  \item With set parameters, we show how to perform an independent selection of multiple objects at once.
\end{itemize}


The rest of the article is organized as follows. Section~\ref{sec:problem} describes the class of studied problems and their state and action spaces. The key elements of our method are described in Section~\ref{sec:method}, i.e., the input processing through a GNN, policy decomposition and training. Section~\ref{sec:experiments} discusses the experiments in three different domains (BlockWorld, Sokoban and SysAdmin), along with their domain definitions and results. In Section~\ref{sec:relatedwork}, the related work is overviewed. Section~\ref{sec:discussion} discusses the architectural choices, the principles of domain modeling and the source of generalization. The article concludes in Section~\ref{sec:conclusion} along with a discussion of future work.

\section{Problem} \label{sec:problem}
Our problems naturally consist of \emph{objects} described with features, heterogeneous binary \emph{relations}, a feature vector describing the \emph{global context}\footnote{The global context, objects and their relations can be viewed as nullary, unary and binary relations.}, and a \emph{goal} definition. The problems are sequential and we assume existing \emph{transition} dynamics. We use the Markov Decision Process (MDP) formalism, i.e., the MDP is a tuple $(\mathcal S, \mathcal A, r, t, \gamma)$, where $\mathcal S$, $\mathcal A$ represent the state and action spaces, $r$, $t$ are reward and transition functions, and $\gamma$ is the discount factor.
All of the MDP components are problem-dependent, hence we provide only their general descriptions. The reward function $r$ can directly specify the goal or be more subtle. For example, in the BlockWorld, the reward can be defined as a small negative value per step and a non-negative value when the goal is reached. The parameter $\gamma$ and the transition function $t$ are directly defined by the particular environment. Because we use a model-free method, the transition function can also be unknown (only a simulator is needed). State and action spaces are described below. 

\subsection{State and goal} 
The state contains objects with their features, relations, global context, and optionally the goal. The objects and relations naturally form an oriented graph where nodes represent the objects and contain their features in the form of fixed-length vectors. More complicated feature structures can be embedded using the existing techniques (e.g., using HMIL \citep{pevny2016discriminative} or Deep Sets \citep{zaheer2017deep}). Heterogeneous objects can be recognized by a type-specifying feature. Oriented edges represent the relations, optionally also containing features and their type. Symmetric relations can be transformed into two opposite edges.
The global context is a vector specifying properties of the environment, unrelated to any single object (e.g., time or the environment state). 

The goal can be encoded in several ways. First, the reward function definition can encode the goal in domains where it is static. In domains where the goal changes across problem instances, it needs to be included in the state. Depending on the particular domain, it can be encoded either in the global context, in the object features, or as part of the graph itself. In the last case, the goal can represent the desired final configuration, encoded as a separate graph and then be joined with the original state.
  
\begin{figure}[t]
  \centering
  \includegraphics[scale=0.6]{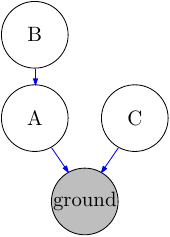} \hfil
  \includegraphics[scale=0.6]{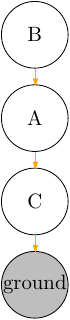} \hfil
  \includegraphics[scale=0.6]{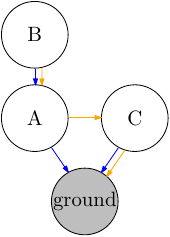}

  (a) current state \hfil (b) goal \hfil (c) combined

  \caption{Example state encoding of the BlockWorld game state from Figure~\ref{fig:domains}a. The objects and relations form the graph, with a special node representing the ground. The representation of the current state and the goal is combined with different edge types. Nodes include a single feature differentiating the blocks and ground; no other features (e.g., labels) are present. Likewise, no global information is needed in this example.}
  \label{fig:bwstate}
\end{figure}

Let us take the BlockWorld domain as an example of the state encoding (see Figure~\ref{fig:bwstate}). Objects and relations are encoded as a graph, with a special node representing the ground (see Sec.~\ref{sec:blockworld} for the domain definition). Nodes contain only a single feature, differentiating between a regular node and the ground. The actual state and the goal are encoded with different edge types, and their representation is combined.

\subsection{Actions}
The actions are object-centric, i.e., they manipulate the problem's objects. It follows that in problems where the number of objects changes, the action space also changes. However, as we show later in Section~\ref{sec:policy_decomposition}, this variable and unbound action space can be tackled efficiently with a fixed-size model.

Let us describe what actions are. They consist of an \emph{action identifier} (e.g., \emph{move}) and their \emph{parameters} (target objects) and \emph{preconditions}. Actions without any parameters are called \emph{elementary} (e.g., \emph{turn-left}). We assume that the parameters are conditionally dependent and need to be selected in a specific order. For example, in the action \emph{move}$(x, y)$, the choice of $y$ depends on the chosen $x$. Moreover, we introduce \emph{set} parameters which are created with any subset of objects (e.g., \emph{select}$(\mathcal X)$, where $\mathcal X$ is an arbitrary set of nodes). The set parameters assume that the nodes can be chosen simultaneously and independently. An action is available only if all its preconditions in a particular state are met. For instance, the \emph{move} action in BlockWorld has two preconditions -- that there is no block on top of $x$, and neither on $y$ (unless $y$ is the ground). 

\section{Method} \label{sec:method}
This section describes the key elements of our method. When reading the following parts, refer to Algorithms~\ref{alg:model} and \ref{alg:training} with a pseudo-code. Our method uses a GNN \citep{zhou2020graph} to process the complex state. To tackle the multi-parameter actions, we use auto-regressive policy decomposition \citep{vinyals2017starcraft}. Our model is fully differentiable and can be trained by any policy gradient algorithm \citep{li2018deep}.

\begin{algorithm}[t]
\caption{SR-DRL: Model}
\label{alg:model}
\begin{algorithmic}[1]
\Function{Model}{state $s$} \Comment{State $s$ includes nodes $\mathcal V$, edges $\mathcal E$, and the initial global context $g$.}
  \State $(\mathcal V, \mathcal E, g) = s$ \Comment{Note that $\mathcal V, \mathcal E, g$ represent both the graph elements and their feature vectors.}
  \State optional: $\forall v \in \mathcal V: v = \phi_{emb\_v}(v); \forall e \in \mathcal E: e = \phi_{emb\_e}(e); g = \phi_{emb\_g}(g)$ \Comment{Embed the features.}
  \For{$l = 1..mp\_steps$} 
    \State $\mathcal V, g$ = \Call{GNN\_MessagePass}{$\mathcal V, \mathcal E, g, l$}
  \EndFor
  \State $a, a_p$ = \Call{SelectAction}{$\mathcal V, g$}
  \State \Return $a, a_p, V_\theta(g)$ \Comment{Return the action, its probability and state value.}
\EndFunction
\item[]

\Function{GNN\_MessagePass}{nodes $\mathcal V$, edges $\mathcal E$, global embedding $g$, level $l$}
  \ForAll {$v \in \mathcal V$}
    \State $\displaystyle v_{msg} = \max_{e \in \mathcal E : e.r=v} \phi_{msg}^{l}(e, e.s)$ \Comment{Aggregate incoming messages; eq.~\eqref{eq:aggregation}.}
    \item[] \Comment{($e.r, e.s$ are the receiving and sending nodes of the edge $e$)}
    \State $v = v + \phi_{agg}^{l}( v, v_{msg}, g )$ \Comment{Update node features; eq.~\eqref{eq:node_update}.}
  \EndFor
  \State $g = g + \phi_{glb}^{l} \Big(g,  \sum_{v \in \mathcal V} \phi_{att}^{l}(v) \cdot \phi_{feat}^{l}(v) ) \Big)$ \Comment{Update the global node; eq.~\eqref{eq:global_update}.}
  \State \Return $\mathcal V, g$
\EndFunction

\item[]
\Function{SelectAction}{$\mathcal V, g$}
  \State Select $a_0$ from $\pi_0(g) = \softmax \phi_{\pi_0}(g)$ \Comment{Select the action id, e.g. \emph{move} or \emph{stop}.}
  \State $a = [a_0]$; $a_p = [\pi_0(a_0 \mid g)]$ \Comment{Store the action and its probability.}

  \For{$l = 1..L(a_0)$} \Comment{Select parameters.}
    \If{$a_l$ should be a normal parameter}
      \State Select $a_l$ from $\pi_{a_0,l}(\mathcal V, g, a_1, ..., a_{l-1}) = \softmax \phi_{\pi_{a_0,l}}(\mathcal V, g, a_1, ..., a_{l-1})$ \Comment{eq.~\eqref{eq:policy_softmax}}
    \ElsIf{$a_l$ should be a set parameter}
      \State Let the probability of choosing a node $v$ be $P(v) = \text{sigmoid}^{(v)} \phi_{\pi_{a_0,l}}(\mathcal V, g, a_1, ..., a_{l-1})$
      \State Create $a_l = \{v_1, v_2, ...\}$ as a set of independently selected nodes with probabilities $P(v)$
      \State Let $\pi_{a_0, l}(a_l \mid \mathcal V, g, a_1, ..., a_{l-1}) = \prod_{v \in a_l} P(v) \cdot \prod_{v \in \mathcal V \setminus a_l} (1-P(v))$. \Comment{eq.~\eqref{eq:policy_set}}
    \EndIf
    \State In case no action $a_l$ is available, disable the $a_{l-1}$ action, go back to the level $l-1$ and re-select.
    \State Append $a_l$ to $a$ and its probability $\pi_{a_0, l}(a_l \mid \mathcal V, g, a_1, ..., a_{l-1})$ to $a_p$ 
  \EndFor
  \State \Return $a, \prod a_p$ \Comment{Return the action with its parameters and the product of the probabilities; eq.~\eqref{eq:policy}.}
\EndFunction

\item[]
\State Let $\phi^{fin}_{\pi_{a_0, l}}: \mathbf R^{|v|+|g|} \rightarrow \mathbf R$ be a linear pre-softmax function.
\State Let $\phi^{emb}_{\pi_{a_0, l}}: \mathbf R^{|v|+l-1} \rightarrow \mathbf R^{|v|}$ be a linear embedding function transforming the augmented vector back to the $\mathbf R^{|v|}$ size.
\Function{$\phi_{\pi_{a_0,l}}$}{$\mathcal V, g ,a_1, ..., a_{l-1}$}
  \If{l = 1}
    \State $\mathcal V', g' = \mathcal V, g$
  \Else
    \State $\forall v \in \mathcal V: z_v \in \{0, 1\}^{l-1}$, where $z_v^{(i)} = 1$ if $a_i = v$, otherwise $0$ \Comment{Create one-hot vectors of past parameters.}
    \State $\mathcal V' = \{ \forall v \in \mathcal V: v' = \Call{LeakyReLU}{\phi^{emb}_{\pi_{a_0, l}}(v, z_v)} \}$ \Comment Concat $v$ and $z$ and transform to the $|v|$ size.
    \State $\mathcal V', g'$ = \Call{GNN\_MessagePass}{$\mathcal V', \mathcal E, g, mp\_steps + 2l - 1$} \Comment{Let the information to spread}
    \State $\mathcal V', g'$ = \Call{GNN\_MessagePass}{$\mathcal V', \mathcal E, g', mp\_steps + 2l$} \Comment{for two steps.}
  \EndIf

  \State Evaluate preconditions given $a_0, ..., a_{l-1}$ and mark possible parameters as $\mathcal{\bar V} \subseteq \mathcal V'$ \Comment{$a_0$ is known from $\pi_{a_0, l}$}
  \State \Return $\left[ \forall v \in \mathcal V': \phi^{fin}_{\pi_{a_0, l}}(v, g) \text{ if } v \in \mathcal{\bar V} \text{ else } -\infty \right]$ \Comment{Return an array of real values.}
\EndFunction
\end{algorithmic}
\end{algorithm}

\begin{algorithm}[t]
\caption{SR-DRL: Training}
\label{alg:training}
\begin{algorithmic}[1]
\Function{Train}{environments $\Sigma$, model $\theta$}
  \State batch $\mathfrak B = [\,]$
  \State $\theta' = \theta$ \Comment{Initialize the target network.}
  \While{not converged}
    \ForAll {$env \in \Sigma$} \Comment{Prepare the batch.}
      \State $s = env.s$ \Comment{We assume the environment provides the state in $\mathcal V, \mathcal E, g$ format.}
      \State $a, a_p, V = \Call{Model}{s}$
      \State $r, s' = \Call{$env$.Step}{a}$ \Comment{We assume a terminal state is returned on an episode finish and the $env$ resets.}
      \State append $s, a, a_p, r, s', V$ to batch $\mathfrak B$
    \EndFor
  \State $L_{pg} = \Call{A2C}{\mathfrak B}$
  \State Update $\theta$ with $\nabla_\theta L_{pg}$
  \State Update the target network: $\theta' := (1-\rho) \theta' + \rho \theta$ 

  \EndWhile
\EndFunction

\item[]
\Function{A2C}{batch $\mathfrak B$} \Comment{with target network and sampled and normalized entropy}
  \State $\nabla_\theta J = \expect_{\mathfrak B} \Big[ (r + \gamma V_{\theta'}(s') - V) \cdot \nabla_\theta \log a_p \Big]$ \Comment{eq.~\eqref{eq:a2c_pg}; $a_p = \pi_\theta(a \mid s)$; $V_{\theta'}(s')$ requires a call to \Call{Model}{}}
  
  \State $L_V = \expect_{\mathfrak B} \Big[r + \gamma V_{\theta'}(s') - V \Big]^2 $ \Comment{eq.~\eqref{eq:a2c_lv}; $V_\theta'(s) = 0$ if $s' = \mathcal T$}

  \State $\nabla_\theta L_H = \expect_{\mathfrak B} \Big[\log a_p \cdot \nabla_\theta \log a_p \Big] $ \Comment{eqs.~\eqref{eq:a2c_lh}, \eqref{eq:a2c_lh2}}

  \State \Return $L_{pg} = -J + \alpha_v L_V - \alpha_h L_H$ \Comment{using auto-differentiation}
\EndFunction
\end{algorithmic}
\end{algorithm}

\subsection{Graph Neural Network}
This section describes the processing of the input state. The pseudo-code is given in Alg.~\ref{alg:model}, methods \Call{Model}{} and \Call{GNN\_MessagePass}{}. Because the state is represented as a graph, the natural choice is to use GNNs, which have strong relational inductive biases \citep{battaglia2018relational} and their operations are local and invariant to node permutations. Also, the same model can be used to process states with a different number of objects. Several GNN variations exist, with a unifying framework made by \citet{battaglia2018relational}. We use a custom implementation that includes node and edge features, skip connections, a global node with an attention mechanism, and separate parameters for each message-passing step. The GNN accepts the state graph $(\mathcal V, g, \mathcal E)$, where $\mathcal V$ are nodes, $\mathcal E$ are oriented edges, and $g$ is a special \emph{global} node. If available, $g$ can initially contain the global context. Optionally, before processing with the GNN, the node, edge and global features can be passed through embedding functions, implemented as a non-linear layer. Several message-passing steps are performed, and the final embeddings are saved in $\mathcal V$ and $g$. 

Let us define the process formally. The input graph consists of nodes $v \in \mathcal V$ and edges $e \in \mathcal E$, where $e.s, e.r$ denote the sending and receiving nodes of this edge. Let $g$ be a special \emph{global} node not included in $\mathcal V$. For simplicity, let $v, g$ and $e$ also denote the feature vector of the respective node or edge.

The core of the algorithm is a single message-passing step. First, the incoming messages are aggregated: 
\begin{equation} \label{eq:aggregation}
  \forall v: v_{msg} = \max_{e \in \mathcal E : e.r=v} \phi_{msg}(e, e.s)
\end{equation}
Here, $\phi_{msg}$ is a message embedding function that transforms an incoming message from node $e.s$ over an edge $e$. The results are aggregated with an element-wise \emph{max} function. Another common aggregation operator is \emph{mean} \citep{battaglia2018relational}; we chose \emph{max} early in our experiments, where it worked best.
Second, all node features are updated with newly computed values:
\begin{equation} \label{eq:node_update}
  \forall v: v' = v + \phi_{agg}( v, v_{msg}, g )
\end{equation}
The messages $v_{msg}$ are processed with the function $\phi_{agg}$, which also takes the current embedding of $v$ and the global node features $g$. In practice, we implement the $\phi_{msg}$ and $\phi_{agg}$ functions as single non-linear neural network layers. The addition of the original $v$ represents a skip connection \citep{kipf2016semi,he2016deep}, which we found to facilitate learning if the number of message-passing steps is large.
After all node representations are updated, a global node $g$ aggregates information from all other nodes through an attention mechanism:
\begin{equation} \label{eq:global_update}
  g' = g + \phi_{glb} \Big(g,  \sum_{v \in \mathcal V} \phi_{att}(v) \cdot \phi_{feat}(v) ) \Big) 
\end{equation}
The $\phi_{att}$ denotes a softmax distribution over all nodes in $\mathcal V$, $\phi_{feat}$ a node embedding function and $\phi_{glb}$ is a final embedding function. In the implementation, $\phi_{att}$ is a single linear layer followed by softmax and $\phi_{feat}$ and $\phi_{glb}$ are single non-linear layers. Again, adding the original $g$ serves as a skip connection to facilitate learning.

The steps \eqref{eq:node_update} and \eqref{eq:global_update} form a single message-passing step, which is repeated $mp\_steps$ times, resulting in final embeddings $v \in \mathcal V$ and $g$. We use independent parameters for the $\phi_{}$ functions for each step \citep[see][]{battaglia2018relational}. This way, the model can compute progressively more complex representations.

\subsection{Policy decomposition} \label{sec:policy_decomposition}
\begin{figure}[t]
  \centering
  \includegraphics[width=0.8\linewidth]{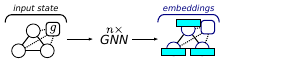} \\
  a) input state embedding \vspace{0.3cm}

  \includegraphics[width=0.8\linewidth]{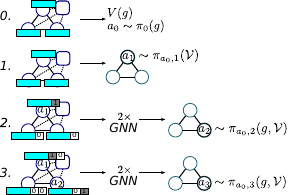} \\
  (b) action selection

  \caption{\textbf{(a)} The input state, including its features, is processed through the GNN, resulting in embeddings of nodes $\mathcal V$ and $g$. \textbf{(b)} The action is selected in a sequence of steps. First, the action identifier $a_0$ (e.g., \emph{move}) is sampled using the embedding of $g$. Next, action parameters $a_1, a_2, ...$ are sequentially chosen (e.g., $x = \mathbf{B}, y = \mathbf{C}$ in the BlockWorld), conditioned on previous selections. Simultaneously, the state value $V(g)$ is computed, which is necessary for the RL algorithm. The $\pi_{a_0, l}$ denotes a policy selecting $l$-th parameter for the action identifier $a_0$.}
  \label{fig:nnstructure}
\end{figure}

This section describes the process of selecting an action. Follow Alg.~\ref{alg:model}, methods \Call{SelectAction}{} and $\phi_{\pi_{a_0,l}}$, and Figure~\ref{fig:nnstructure}. The policy $\pi(s)$ is a probability distribution over all possible actions in a state $s$. Although the action space grows exponentially with the number of actions' parameters, the selection of a particular action can be done in a linear time by decomposing the policy into a sequence of choices. Let $\mathcal A_0$ be the set of action identifiers, e.g., \{\emph{stop}, \emph{move}, ...\}. $L(a)$ is the arity of action $a$ with its parameters, $a=(a_0, a_1, ..., a_{L(a)})$. Here, $a_0 \in \mathcal A_0$ denotes the action identifier and $a_1, a_2, ...$ are the action's parameters selecting graph nodes, $a_{1..|L(a)|} \in \mathcal V$. The policy can then be represented in an auto-regressive manner:
\begin{equation} \label{eq:policy}
  \pi(a \mid s) = \pi_0(a_0 \mid g) \prod_{l=1}^{L(a)} \pi_{a_0,l}(a_l \mid \mathcal V, g, a_1, ..., a_{l-1})
\end{equation}
where $\pi_0$ is the policy selecting the action identifier, and $\pi_{a_0,l}$ is the policy selecting a the action's $l$-th parameter. In practice, an action can be selected by sequentially sampling the policies in eq.~\ref{eq:policy}.

A similar policy decomposition was studied by \citet{vinyals2017starcraft}, who chose to disregard the conditional dependency on the previously chosen parameters. However, this variant cannot represent every possible probability distribution, and for some actions, the previously selected parameters are crucial for further selection. For example, in \emph{move(x,y)}, the selection of \emph{y} only makes sense with a known \emph{x}. Therefore, we propose the following method to preserve the conditional dependency (see Figure~\ref{fig:nnstructure}).

First, the action identifier $a_0$ is selected from $\pi_0(g) = \softmax \phi_{\pi_0}(g)$. We suggest to implement the projection $\phi_{\pi_0}: \mathbf R^{|g|} \rightarrow \mathbf R^{|A_0|}$ as a linear layer.
For elementary actions without parameters, the decision ends here. Otherwise, the action parameters (graph nodes) are selected sequentially and conditioned on previous selections. Specifically, the $l$-th parameter $a_l$ is chosen as:
\begin{equation} \label{eq:policy_softmax}
  a_l \sim \pi_{a_0,l}(\mathcal V, g, a_1, ..., a_{l-1}) = \softmax \phi_{\pi_{a_0,l}}(\mathcal V, g, a_1, ..., a_{l-1})
\end{equation}
Here, the function $\phi_{\pi_{a_0,l}} : \mathbf R^{|\mathcal V| \times (|v|+l-1)+|g|} \rightarrow \mathbf R^{|\mathcal V|}$ transforms the nodes' embeddings, the global embedding and a one-hot encoding of previously selected parameters into one real value per node, used for the softmax (see Alg.~\ref{alg:model}, method $\phi_{\pi_{a_0,l}}$). For the first parameter, $\phi_{\pi_{a_0,1}}$ is simply a linear projection applied uniformly to all nodes.

For further parameters ($l \geq 2$), the computation of $\phi_{\pi_{a_0,l}}$ need to take the previous selection $a_1, ..., a_{l-1}$ into account. To this purpose, each node is augmented with a one-hot encoding of $[a_1, ..., a_{l-1}]$, i.e., every node is augmented with a binary vector of size $l-1$, where $i$-th element is $1$ if the node was selected as an $i$-th parameter, else $0$. To preserve the original embedding size, the augmented vector is projected to its original size and at least two message-passing steps are performed to allow the information to spread. We found that two passes worked well for BlockWorld, but some domains may require more. The result of the operation are alternative $v', g'$, which are used to determine the final real value for each node. 

\subsubsection{Preconditions}
Preconditions determine whether an action is available in a particular situation. The availability is resolved for the currently processed level (e.g., for $a_0$, $a_1$, ...), and the unavailable actions are removed from the softmax computation. In the most general case, no selection may be possible for a particular level $l$. In that case, the algorithm has to backtrack, disable the selection at level $l-1$ that led to the situation and select a new parameter.

\subsubsection{Set parameters}
The set parameters consist of an arbitrary subset of nodes. To perform such selection, we use concurrent actions \citep{harmer2018imitation} (follow Algorithm~\ref{alg:model}, lines 24-27). A shared function with sigmoid activation is used to compute per-node probabilities $P(v)$. Then, nodes are selected with independent Bernoulli trials. Let $a_l$ be the parameter with the set of selected nodes. Its total probability is then:
\begin{equation} \label{eq:policy_set}
 \pi_{a_0, l}(a_l \mid \mathcal V, g, a_1, ..., a_{l-1}) = \prod_{v \in a_l} P(v) \cdot \prod_{v \in \mathcal V \setminus a_l} (1-P(v))
\end{equation}

\subsubsection{Complexity analysis}
In a graph with $n$ nodes and a maximal degree $k$, the time complexity of one message passing step is $O(kn)$. Parameters of an action with $p$ parameters are selected sequentially, and two message passes are performed for each. The time complexity of selecting the action is then $O(pkn)$, while $k$ is usually very small. Information can be sequentially accumulated in nodes, hence space complexity is $O(n)$.

\subsection{Model training}
Here, we describe how the model is trained. See Alg.~\ref{alg:training}, methods \Call{Train}{} and \Call{A2C}{}. Let $\theta$ be the model parameters -- a union for all $\phi$ functions and layers used in the action selection. Apart from the parametrized policy $\pi_\theta$, the model includes a separate output head for a state value estimate. It is implemented as a single linear neural network layer $V_\theta(g)$, taking the final embedding of the global node $g$. Although the action selection involves deterministic choices of their parameters, the final product $\pi_\theta(a|s)$ is fully differentiable. We propose to use A2C algorithm, a synchronous version of A3C \citep{mnih2016asynchronous}, where we included a target network \citep{lillicrap2015continuous} and entropy gradient sampling \citep{zhang2018efficient}. We note the method could be modified for other policy gradient methods (e.g., PPO \citep{schulman2017proximal}). The algorithm works as follows:

Let $\pi_\theta$ be a policy and $V_\theta$ a value estimate, where $\theta$ are model parameters. In a Markov Decision Process (MDP) $(\mathcal S, \mathcal A, r, t, \gamma)$, where $\mathcal S, \mathcal A$ are state and action spaces, $r$, $t$ are reward and transition functions and $\gamma$ is a discount factor, let the state-action function $Q(s, a)$ be:
\[ Q(s, a) = \expect_{s' \sim t(s, a)} \Big[ q(s, a, s') \Big ] \; ; \; %
q(s, a, s') = 
  \begin{dcases*}
    r(s, a, s')                         & if $s'$ is terminal \\
    r(s, a, s') + \gamma V_{\theta'}(s')  & else
  \end{dcases*} \]
To stabilize training, the $V_{\theta'}(s')$ is estimated using the target network with a copy of parameters $\theta'$ that are regularly updated with $\theta' := (1-\rho) \theta' + \rho \theta$, with $\rho \in (0,1]$. Let $A(s, a) = Q(s, a) - V_\theta(s)$ be an advantage function. Then, the policy gradient $\nabla_\theta J$ and the value function loss $L_V$ are:
\begin{equation} \label{eq:a2c_pg}
\nabla_\theta J = \expect_{s, a \sim \pi_\theta, t} \Big[ A(s, a) \cdot \nabla_\theta \log \pi_\theta(a | s) \Big]
\end{equation}
\begin{equation} \label{eq:a2c_lv}
L_V = \expect_{s, a \sim \pi_\theta, t} \Big[ Q(s, a) - V_\theta(s) \Big]^2
\end{equation}
An entropy regularization term $L_H$ is:
\begin{equation} \label{eq:a2c_lh}
L_H = \expect_{s \sim \pi_\theta, t} \Big[ H_{\pi_\theta}(s) \Big] \; ; \; H_{\pi}(s) = -\expect_{a \sim \pi(s)} \Big[ \log \pi(a|s) \Big]
\end{equation}
However, the precise computation of the policy entropy is intractable in our case -- only single $\pi(a|s)$ for the actually performed action $a$ is available. Using a collection of sampled actions in a batch, the gradient of $H_{\pi_\theta}$ can be estimated as \citep{zhang2018efficient}:
\begin{equation} \label{eq:a2c_lh2}
  \nabla_\theta H_{\pi_\theta}(s) = -\expect_{a \sim \pi_\theta(s)} \Big [ \log \pi_\theta(a|s) \cdot \nabla_\theta \log \pi_\theta(a|s) \Big]
\end{equation}

The final gradient is $\nabla_\theta (- J + \alpha_v L_V - \alpha_h L_H)$, with $\alpha_v, \alpha_h$ being learning rate coefficients. We simulate a batch of parallel environments to gather a better gradient estimate and perform a single update per each step of the environment.

\section{Experiments} \label{sec:experiments}
To demonstrate our method's generality and performance, we provide an implementation and experimental results in three distinct domains, each of which represents a different class of problems:
\begin{itemize}
  \item In \textbf{BlockWorld}, we showcase a multi-parameter action and preconditions. Additionally, the graph structure changes during an episode.
  \item \textbf{Sokoban} is a domain with multiple single-parameter actions where long-term planning is required. We design our agent to operate on the object level (boxes), while a low-level planner breaks them into micro-actions (movement of the player). Note that translating the object-level actions to micro-actions is a polynomial problem.
  \item \textbf{SysAdmin} is a graph domain where only short-distance information is usually required to perform well. It includes stochastic transitions and an infinite time horizon, without any specific goal to reach. We include a separate variation of this domain, SysAdmin-M, which represents problems where multiple objects can be independently selected. 
\end{itemize}
In all three domains, we study how a trained model translates to problem instances with different sizes. In the following sections, we study each of the domains separately from different points of view. In each part, we introduce one domain, provide its detailed definition and perform a set of unique experiments.

\subsection{Experiment setup}
Before we start describing the domains, let us introduce the details regarding the implementation and used hardware.

\subsubsection{Time-limits}
The used environments are not restricted by any time horizon, hence we use a discount factor $\gamma=0.99$ in all domains. To enhance the training experience diversity and avoid possible deadlocks (e.g., in Sokoban), we employ an artificial step limit (100 in BlockWorld and SysAdmin, 200 in Sokoban). This limit is regarded as auxiliary and not part of the environment, in the spirit of \citet{pardo2018time}.

\subsubsection{Reference machine} When we report our algorithm's running times in the following text, we are using our reference machine equipped with AMD Ryzen 1900X CPU, 8~GB of RAM, and nVidia Titan X GPU.

\subsubsection{Implementation details}
For all non-linear layers, we use the LeakyReLU activation function \citep{maas2013rectifier}, unless specified otherwise. Before processing the state in the GNN, objects' features are embedded into a fixed-length vector of size $emb\_size$, with a shared single non-linear layer. The same parameter $emb\_size$ then defines the dimension of all subsequent intermediary embeddings of nodes and the global context. Edge types are one-hot-encoded and used directly. Message-passing steps (eqs.~\ref{eq:node_update},~\ref{eq:global_update}) are repeated $mp\_steps$ times to get the final embeddings $\mathcal V, g$.
The AdamW optimizer \citep{loshchilov2017decoupled} with a weight decay of $1\times10^{-4}$ is used. Gradient norm is clipped to $grad\_max\_norm$. The learning rate and the entropy regularization coefficient are annealed from their respective starting values $LR$ and $\alpha_{h}$ until their minimum $\frac{1}{30}LR, \frac{1}{2}\alpha_{h}$. The learning rate annealing schedule is step-based, with a factor $0.5$ used every $20 \times epoch$ steps. The coefficient $\alpha_h$ is annealed using a $\frac{1}{t}$ schedule, where $t$ is increased per each $epoch$ steps. For each environment, we define a $q\_range$ interval, that is used to clip the target $Q(s, a)$ in eq.~\ref{eq:a2c_lv}. A batch of $p\_envs$ environments is simulated in parallel. Many of the parameters were found using a grid-search in their respective domains. Used resources and other hyper-parameters are available in Table~\ref{tab:hyperparameters}.

\begin{table}
    \caption{Hyper-parameters and other settings used in the experiments}
    \begin{tabular}{lrrr}
      \toprule
      parameter & BlockWorld & Sokoban & SysAdmin-S/-M \\
      \midrule                                
      $p\_envs$, batch size & 256 & 256 & 256 \\
      $\rho$, target-network update coefficient & 0.005 & 0.005 & 0.005\\
      $\gamma$, discount factor & 0.99 & 0.99 & 0.99 \\
      $epoch$, number of steps per epoch & 1000 & 1000 & 100 \\
      episode step-limit & 100 & 200 & 100 \\
      $mp\_steps$ number of message-passes & 3 & 10 & 5 \\ 
      $emb\_size$, embedding size & 32 & 64 &  32 \\
      $LR$, initial learning rate & $3\times10^{-4}$ & $3\times10^{-3}$ & $3\times10^{-3}$ \\
      $grad\_max\_norm$, maximal gradient & 3.0 & 5.0 & 3.0 \\
      $q\_range$, range of $Q(s, a)$ (eq.~\ref{eq:a2c_lv}) & $[-15, 15]$ & $[-15, 15]$ & $[-100, 200 \cdot N]$ \\
      $\alpha_v$, coefficient of $L_V$ & 0.1 & 0.1 & 0.1 \\
      $\alpha_{h}$, coefficient of $L_H$ & $2.5\times10^{-5}$ & 0.04 & $0.15 \sim 0.5$ (-S)\footnotemark[1] \\
          & & &                                                        $0.1 \sim 0.2$ (-M) \\


      \midrule
      resources used in training & 4 CPU cores & 2 CPU cores, 1 GPU & 1 CPU core \\
      \bottomrule
    \end{tabular}      
    \footnotetext[1]{$0.15/0.15/0.3/0.3/0.5/0.5$ in -S, $0.1/0.1/0.2/0.2/0.2/0.2$ in -M for $N = 5/10/20/40/80/160$}

    \label{tab:hyperparameters}
\end{table}
\subsection{BlockWorld} \label{sec:blockworld}
BlockWorld is a well-known domain with tractable satisficing planning and NP-hard optimal planning \citep{slaney2001blocks}. 
This environment consists of $N$ blocks and a special ground object. The blocks can be placed on top of each other or on the ground. A single \emph{move}$(x, y)$ action with two parameters is available, which picks a block $x$ and puts it on top of $y$. Its preconditions are that $x$ and $y$ are free, unless $y$ is the ground. Note that we use the \emph{move} action with two parameters deliberately to demonstrate that our framework works with multi-parameter actions. In planning, the BlockWorld domain is usually defined with single-parameter actions \emph{pickup(x)} and \emph{putdown-to(y)}, which makes the problem easier. The goal is to reconfigure the blocks from a starting position to a given goal position. The agent receives a small penalty per step and a reward for solving the problem. 

\subsubsection{Detailed domain definition}
\paragraph{Definition.}
The objects in the BlockWorld problem consist of a set of $N$ blocks $\mathcal B = \{b_1, b_2, ..., b_N\}$ and a special object $G$, representing the ground. Let's define a relation $x \dashv y; x \in \mathcal B, y \in {\mathcal B \cup G}$, meaning that a block $x$ is positioned on top of $y$. For each $x$, the relation is unique, as well as for each $y$, unless $y = G$. Let $\mathcal R$ be a set of all relations in the problem. The action $move(x,y)$ removes all relations $x \dashv z; \forall z$ from $\mathcal R$ and creates a new one $x \dashv y$. The preconditions for the action are $x \neq y$, $free(x)$ and $free(y) \lor y = G$, where $free(x) \Leftrightarrow \nexists z: z \dashv x$.

The goal is to use the action $move$ to reconfigure the block positions $\mathcal R_{start}$ into $\mathcal R_{goal}$. To incentivize the agent to find the optimal solution, it receives a reward $-0.1$ for each action. After reaching the goal, the episode ends with a reward $10$.

\paragraph{State and actions.}
The state consists of the objects $\mathcal B, G$, the current set of relations $\mathcal R$, and the goal $\mathcal R_{goal}$ (see Figure~\ref{fig:bwstate} for illustration). In the graph, each relation is modeled symmetrically (both \emph{above-of} and \emph{below-of} are included). The different types of relations are marked by their edge parameters. The objects contain a single-bit feature that signifies whether they belong to $\mathcal B$ or $G$. Note that no block labels are present in the state in any way.

There is a single action $move$ with two parameters. The preconditions are used according to their definition. If a particular block is allowed to be the first parameter of the action $move$, there always exists a valid second parameter (e.g., $G$).

\paragraph{Generation.}
A problem instance is generated as follows. From a set of $N$ available blocks $\mathcal B' = {b_1, ..., b_N}$ a random subset of $1$ to $|\mathcal B'|$ blocks is chosen and stacked in random order. This stack is then removed from $\mathcal B'$, and the procedure repeats until $\mathcal B'$ is empty. The goal is generated in the same way.  

\begin{figure}[t]
  \centering
  \includegraphics[width=\linewidth]{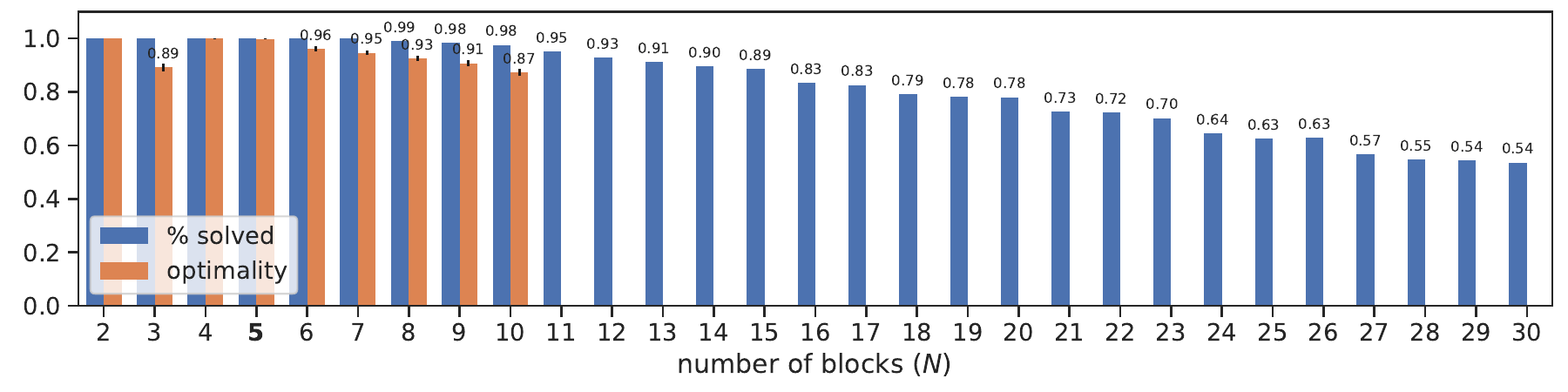} 

  \caption{The agent \textbf{trained} in BlockWorld \textbf{with 5 blocks} ($N=5$) and evaluated in problems with $N \in \{2..30\}$. For each $N$, the agent is evaluated in 1000 problems, and the percentage of solved problems and its optimality (optimal / performed steps) for $N \leq 10$ is reported.}
  \label{fig:blockworld-generalization}
\end{figure}
\begin{figure}[t]
  \centering
  \includegraphics[width=\linewidth]{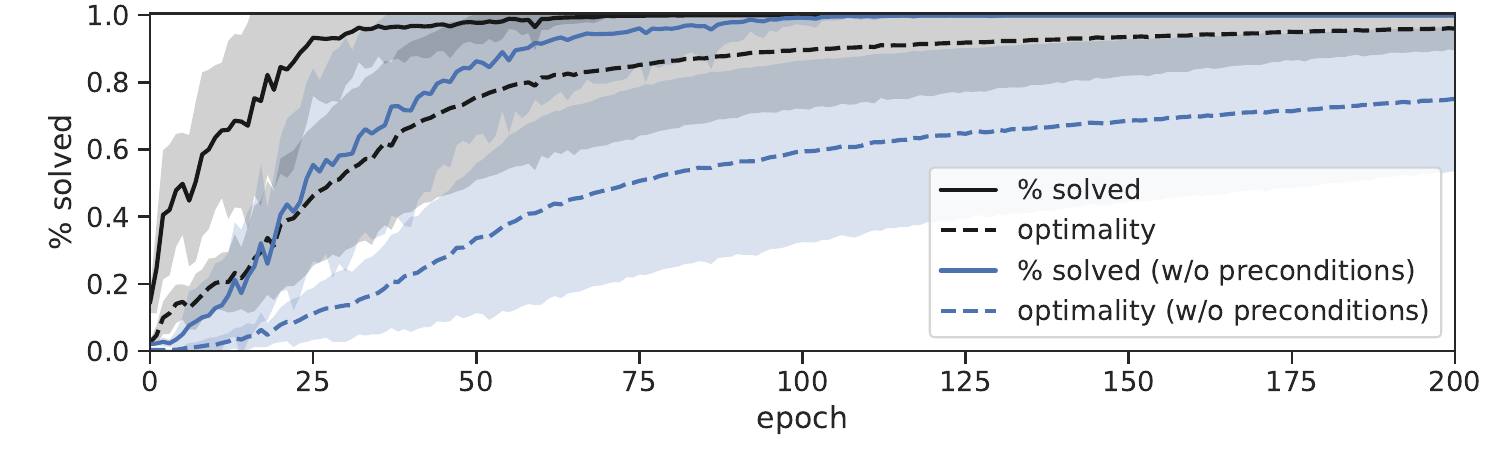} 

  \caption{Training in the BlockWorld environment, $N=5$, with and without preconditions. The graph shows the percentage of solved problems and optimality; each epoch equals 256k environment steps. The mean ± one standard deviation of eight randomly initialized runs is displayed.}
  \label{fig:blockworld-perf}
\end{figure}

\subsubsection{Primary results}
We trained eight models with different seeds in the BlockWorld environment with $N=5$ and randomly generated initial states and goals. The agent is evaluated on 1000 random problems with $N=5$, and we report the percentage of solved problems and optimality -- the average ratio of the number of optimal steps and performed steps for each problem. In case the agent does not solve the environment in the 100 step limit, we consider the ratio to be 0. Figure~\ref{fig:blockworld-perf} shows the first 200 epochs, where a single epoch is 256k environment steps (1000 gradient updates with a batch size of 256). On our reference machine, a single epoch takes about 3.3 minutes; 100 epochs take about 5.5 hours.
We measured the number of optimal steps using Fast Downward planner \citep{helmert2006fast} with A* algorithm and LM-cut heuristic \citep{helmert2011lm}.

At about epoch 75, the agent learns to solve the problems with 100\% accuracy and  83\% optimality. Subsequently, the optimality increases; in epoch 200 it's 96\%, and it reaches 99\% in epoch 400. Hence, it can be said that the agent is able to learn near-optimal policy in this setting.

Next, we investigated a variant without preconditions. In this experiment, we enabled all actions and let the agent learn to ignore the nonsensical ones. The results show the training time is almost doubled (see Fig.~\ref{fig:blockworld-perf}). In this case, the agent solves 100\% of problems at about epoch 130 and reaches 99\% optimality at about epoch 730. We conclude that the preconditions are not necessary, but greatly help the training. 

\subsubsection{Additional experiments}
Next, we focused on the agent's generalization to a different number of blocks. From the eight runs with $N=5$, we picked the one that performed best after 800 epochs. Next, we evaluated it in environments with a different number of blocks, $N \in \{2..30\}$. Again, we measured the percentage of solved environments (within the 100 step limit) and the agent's optimality. Because in BlockWorld, the optimal planning is NP-hard, we report the optimality only for $N \leq 10$; it becomes too expensive to compute with higher $N$.
The results, reported in Figure~\ref{fig:blockworld-generalization}, show impressive generalization capabilities. The agent zero-shot generalizes with great success to other problem sizes. It solves all problems for $N \leq 5$ with near-100\% optimality, with the exception of $N=3$. With $N \geq 6$, the fraction of solved problems gradually decreases to 78\% for $N=20$ and 54\% for $N=30$. The optimality decreases to 87\% for $N=10$. 

We tried to train an agent directly with $N=10$, but the agent did not learn any useful policy. The reason is that it is hard to find a solution in a large state space. With ten blocks, there are about $10^7$ box combinations and the only positive reward is received when the problem is solved.  Yet, the agent trained with $N=5$ is able to solve 98\% of problems with $N=10$, with 87\% optimality. Moreover, it gracefully generalizes up to $N \leq 30$, possibly even more. This indicates a strong potential for curriculum learning \citep{bengio2009curriculum}. To understand how impressive these results are, we note that the number of all possible block configurations rises very quickly with $N$. Exactly, it is $\sum_{i=0}^{N} \binom{N}{i} \frac{(N-1)!}{(i-1)!}$ \citep{slaney2001blocks}; i.e., 501 for $N=5$, $5.8\times10^7$ for $N=10$ and $2.7\times10^{20}$ for $N=20$. The number of actions is $|\mathcal A| \leq N^2$.

\subsection{Sokoban}
\begin{figure}[t]
  \centering
  \setlength\tabcolsep{4pt}
  \begin{tabular}{lcccc}
    &
    \includegraphics[scale=0.4]{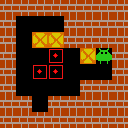} &
    \includegraphics[scale=0.4]{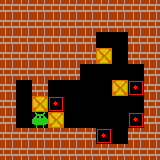} &
    \includegraphics[scale=0.4]{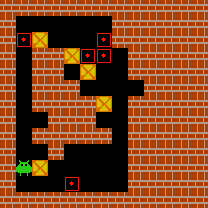} &
    \includegraphics[scale=0.4]{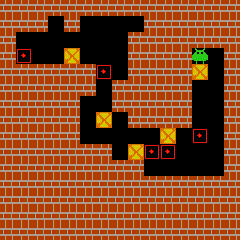} \\

    & 8×8 & \textbf{10×10} & 13×13 & 15×15 \\
    & 3 boxes & \textbf{4 boxes} & 5 boxes & 5 boxes \\
    \midrule
    solved & 95.7\% & 95\% & 84.6\% & 83.9\% \\ 
    \bottomrule
  \end{tabular}

  \caption{We evaluated an agent \textbf{trained solely on the 10×10 levels with 4 boxes} on randomly generated levels in other game variations and measured the percentage of solved levels. The agent generalizes well to different game sizes and the box count. The results show the performance of a single agent evaluated in about 2000 problems per variation. The top row shows example levels.}
  \label{fig:sokoban-envs}
\end{figure}

Sokoban (see Figure~\ref{fig:sokoban-envs}) is a classic planning domain, where an agent moves inside a grid maze with the goal of pushing boxes onto their destination. Solving levels requires careful planning because some actions are irreversible and can lead to an unsolvable situation. Usually, the actions control the player avatar and are elementary -- \emph{left}, \emph{up}, \emph{right}, and \emph{down}. However, our framework's strength lies in its ability to directly manipulate objects. Hence, we define new \textbf{actions that operate directly on the boxes}, with a low-level planner that trivially translates them into the elementary actions while preserving all the environment's mechanics. These new actions are \emph{push-left}$(x)$, \emph{push-right}$(x)$, \emph{push-up}$(x)$, and \emph{push-down}$(x)$, all of which operate on a box $x$, moving the player such that the box is pushed to the desired direction, if possible. Finding the low-level plan with elementary actions corresponding to the \emph{push-} macro actions is trivial, because it can be found in polynomial time.

Note that this abstraction is natural in our framework, which treats all boxes the same and does not need their identifiers, nor absolute positions. The benefit of our method is that the macro actions generalize over the boxes by using the same parameters for each box. Also, it naturally scales to any number of boxes. If we wanted to use the same abstraction with traditional Deep RL with a fixed number of actions, it is possible for a fixed number of boxes (e.g., an agent trained for 4 boxes would use $4 \times 4=16$ actions). The traditional Deep RL would learn separate parameters for each action and it would not work with other number of boxes than what it was trained with.

\begin{figure}[t]
  \centering
  \includegraphics[width=0.49\linewidth]{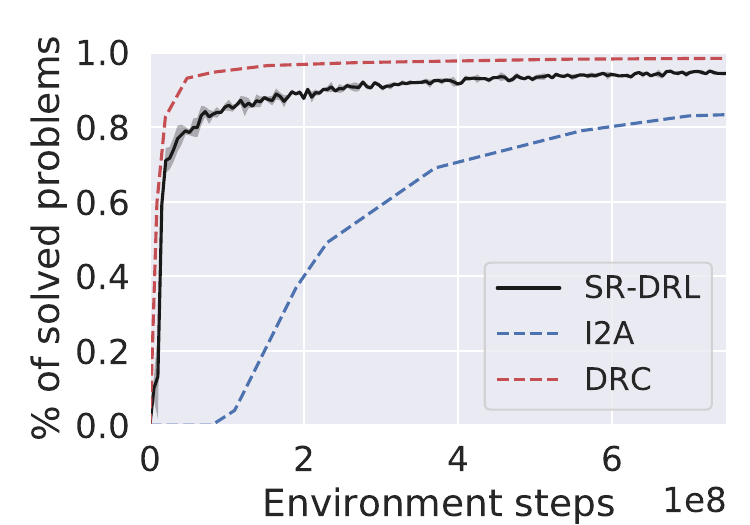} %
  \includegraphics[width=0.49\linewidth]{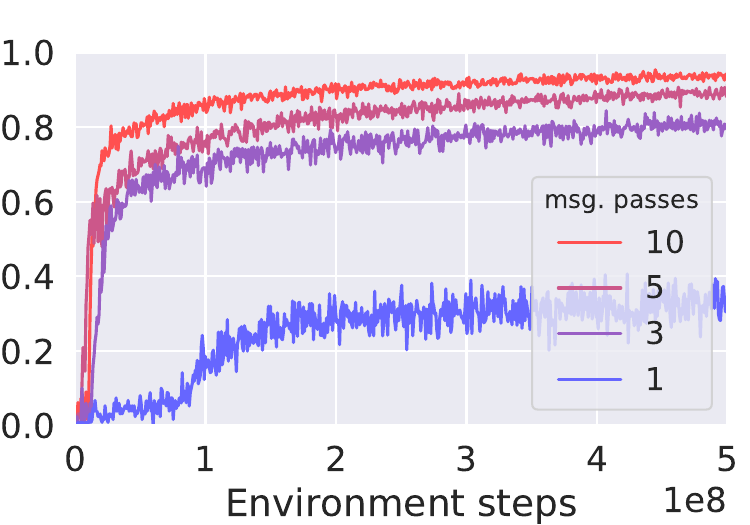} 

  \caption{\emph{Left:} Test-set performance during training on 10×10 levels with 4 boxes. SR-DRL shows a mean ± one standard deviation of three models. I2A and DRC are different Deep RL-based algorithms, without the ability to transfer to larger problem instances. \emph{Right:} Number of message-passes greatly influences the model's performance. With more than 10 message passes, the model failed to train.}
  \label{fig:sokoban-perf}
\end{figure}

\subsubsection{Detailed domain definition}
\paragraph{Definition.}
We use the Sokoban environment as defined in \citet{racaniere2017imagination}. For our purposes, a Sokoban problem is determined by four matrices $W, G, B, P$ with sizes corresponding to the problem size. Here, $W_{xy}=1$ if there is a \emph{wall} at position $x, y$, else 0. Similarly, $G$ determines the position of \emph{goals}, $B$ \emph{boxes}, and $P$ the \emph{player}. The agent can perform five actions \emph{left}, \emph{right}, \emph{up}, \emph{down}, and \emph{no-op} (note that these are the low-level actions defined by the environment, but the SR-DRL agent actually works with different, high-level actions specified below). These low-level actions move the player in the specified direction, possibly pushing a box, and modify the matrices $B, P$ accordingly. The \emph{no-op} action does nothing. The problem is solved when all boxes are on the goal positions, i.e., $B=G$. 

For each action, the agent receives a penalty $-0.1$, plus a reward $1$ if it pushes a box on a goal, $-1$ if it pushes a box off a goal, and $10$ when the level is solved.

\paragraph{State and actions.}
A state is defined as a graph with a node $n_{xy}$ for every $W_{xy}=0$, i.e., only the playable spaces without walls. Compared to convolutional neural networks (CNN) that end with a fully connected layer, our architecture is scale-invariant, accommodates to the particular problem and can save computational resources. The features of every $n_{xy}$ node are defined as a concatenation of $G_{xy}, B_{xy}$ and $P_{xy}$. We tried including positional $x, y$ features, but found no difference in performance. Every two neighboring nodes (in the four directions in the grid) are connected with two opposite edges. Each edge contains a single feature determining its original direction (up, down, left, right).

Rather than using the elementary actions of the environment, we take advantage of the unique ability of our framework to define actions that operate directly on the available objects. We define the following macro actions: \emph{push-left}$(x)$, \emph{push-right}$(x)$, \emph{push-up}$(x)$, \emph{push-down}$(x)$. These actions operate directly on the boxes at the place of node $x$ -- the player walks to a proper spot and pushes a box in the corresponding direction (if possible). Preconditions force the actions to select a node with a box. All actions use a simple A*-based planner that maps them onto the elementary actions of the environment. It may happen that an action is not executable because a path to the right location does not exist. In these cases, the \emph{no-op} action is performed instead. The reward is defined as a sum of rewards resulting from the execution of the related low-level actions. 

\paragraph{Generation.}
We used an implementation of Sokoban provided by \citet{schrader2018sokoban} and the \emph{unfiltered} dataset from \citet{guez2018boxoban}, which contains 900k pre-generated levels of size 10×10 with 4 boxes. For randomly generated levels, we use a method described by \citet{racaniere2017imagination}.

\subsubsection{Primary results}
We trained three models with a dataset of 10×10 problems with four boxes over the course of 17 days. Figure~\ref{fig:sokoban-perf}-left shows the test set performance measured during the training (elementary steps are used). We also show the performance of two other Deep RL-based architectures: I2A \citep{racaniere2017imagination} and DRC \citep{guez2019investigation}, as reported in the original papers (both were trained with the same dataset).
I2A is based on convolutional neural networks (CNN) and learns an environment model used to simulate trajectories; I2A with 15 unrolls is reported. DRC is a recurrent CNN-based architecture; the DRC(3,3) version is reported.

After $10^9$ elementary environment steps, SR-DRL solves 96\% of test levels. In the same amount of steps, I2A reaches 90\% solved levels, and DRC 99\%. We hypothesize that the main advantage of DRC architecture is in its recurrence, allowing it to store intermediary calculations between steps and thus be much more effective. Recurrent architecture can also be used with our method and is a promising future direction. Although DRC outperforms our method in this case, it cannot generalize to different problem sizes. Its CNN architecture ends with a fully connected layer, fixing it to the particular problem size. 

On the other hand, our method is not fixed to any problem size. In Figure~\ref{fig:sokoban-envs}, we report results obtained by evaluating a model trained in 10×10 problems with 4 boxes on several different problem sizes. The results are impressive -- the model generalizes well to both smaller and larger environments, e.g. in 8×8 with 3 boxes it solves 96.7\% of problems, in 15×15 with 5 boxes it is 89\%.

\subsubsection{Additional experiments}
In a separate experiment, we tested the influence of the number of message-passing steps. Figure~\ref{fig:sokoban-perf}-right shows the training progress with 1, 3, 5 and 10 message passes. More message passes always result in faster learning and better final performance. However, when we further increased the number of message passes to 15 or 20, the model did not train at all and the final performance was zero.

Finally, we performed a similar experiment as in BlockWorld -- we compared the learning process with and without the preconditions. In Sokoban, the preconditions mask out the nodes without any boxes, hence forcing the \emph{push-} actions to select a place with a box. However, in this environment, we found that disabling preconditions did not result in any degradation of the performance. Apparently, the model trained without preconditions learns to ignore the meaningless actions early in the training.

\subsection{SysAdmin}

\begin{figure}[t]
  \centering
  \includegraphics[width=0.55\linewidth]{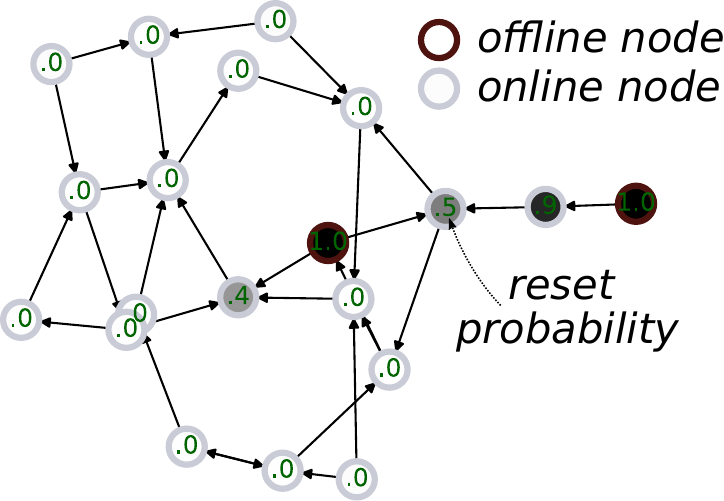} 

  \caption{In SysAdmin-M, SR-DRL learns to preventively reset nodes that have a high probability of failure. The graph shows the dependency network and the reset probabilities the algorithm assigns to the nodes. Read any edge $(a,b)$ as $b$ depends on $a$.}
  \label{fig:sysadmin-deps}
\end{figure}

SysAdmin \citep{guestrin2003efficient} (see Figure~\ref{fig:sysadmin-deps}) is a probabilistic planning domain adapted from the International Probabilistic Planning Competition (IPPC) 2011. It includes stochastic transitions and an infinite time horizon, without any specific goal to reach. In this domain, the problem is defined as a graph of dependencies between $N$ computer nodes. At each step, any computer can be either \emph{online} or \emph{offline}. Online computers have a certain probability of becoming offline, based on the state of their dependencies, and offline computers have a chance to spontaneously reset and become online. We investigate two variants of the problem: In \emph{SysAdmin-S} (\emph{S} for \underline{S}ingle), the agent can perform a \emph{reset}$(x)$ action, that resets a single computer. In \emph{SysAdmin-M} (\emph{M} for \underline{M}ulti), the action \emph{reset}$(\mathcal X)$ contains a set parameter $\mathcal X$ and it reboots an arbitrary set of computers at once. At each step, the agent is rewarded for each computer that is online at that moment and penalized for any computer it reboots.

\subsubsection{Detailed domain definition}
\paragraph{Definition.}
In the SysAdmin domain, an oriented graph represents a computer network. The nodes represent the computers $\mathcal C=\{c_1, c_2, ..., c_N\}$ and each edge $(c_i, c_j) \in \mathcal E$ represents a dependency of $c_j$ on $c_i$. At each timestep $t$, each computer can be either \emph{online} or \emph{offline}. Let $on_t(c) = 1$ if $c$ is online at step $t$, else it is $0$. Let $D(c_j) = \{c_i : (c_i, c_j) \in \mathcal E\}$ be a set of all computers that $c_j$ depends on and let $D_{on_t}(c_j) = \{c_i : (c_i, c_j) \in \mathcal E \land on_t(c_i) \} $ be a set of computers that $c_j$ depends on and which are online at step $t$. At each step, computers that are online have a chance to shut down and offline computers have a chance to reboot and become online. Without any intervention, the network evolves as follows:
\[ P \Big( on_{t+1}(c) = 1 \Big) = 
  \begin{dcases*}
    0.9 \cdot \frac{1 + |D_{on_t}(c)|}{1 + |D(c)|}  & if $on_t(c) = 1$ \\
    0.04                                            & if $on_t(c) = 0$
  \end{dcases*}
 \]

At each step $t$, an action $reset(\mathcal X_t)$ can be performed. It resets the targeted computers $\mathcal X_t$, such that $on_{t+1}(c) = 1 : \forall c \in \mathcal X_t$. Note that it may be reasonable to reset online nodes to make sure they do not go offline at the next step. At each timestep, the agent receives a reward:
\[ r_t =  \sum_{c \in C} on_t(c) - 0.75 \cdot |\mathcal X_t| \]

We investigate two variants of the problem: In \emph{SysAdmin-S} (\emph{S} for \underline{S}ingle), only a single computer can be selected, $|\mathcal X_t| \leq 1$. In \emph{SysAdmin-M} (\emph{M} for \underline{M}ulti), an arbitrary set of computers can be reset.

\paragraph{State and actions.}
The graph of computers $\mathcal C$ and static dependencies $\mathcal E$ constitute the state. Each computer $c$ has a single bit feature $on_t(c)$, determining whether it is online. The edge orientations represent the dependencies. In SysAdmin-S, two actions are available: $noop$ and $reset(c)$. The first action does not select any computer to restart, the second action selects one. In SysAdmin-M, there is only one action $reset(\mathcal X)$ with a set parameter $\mathcal X$, which allows to select an arbitrary set of computers.

\paragraph{Generation.}
For each node of a graph with $N$ nodes, from 1 to 3 (uniformly chosen) other nodes become its dependees. 

\begin{figure}[t]
  \centering
  \includegraphics[width=\linewidth]{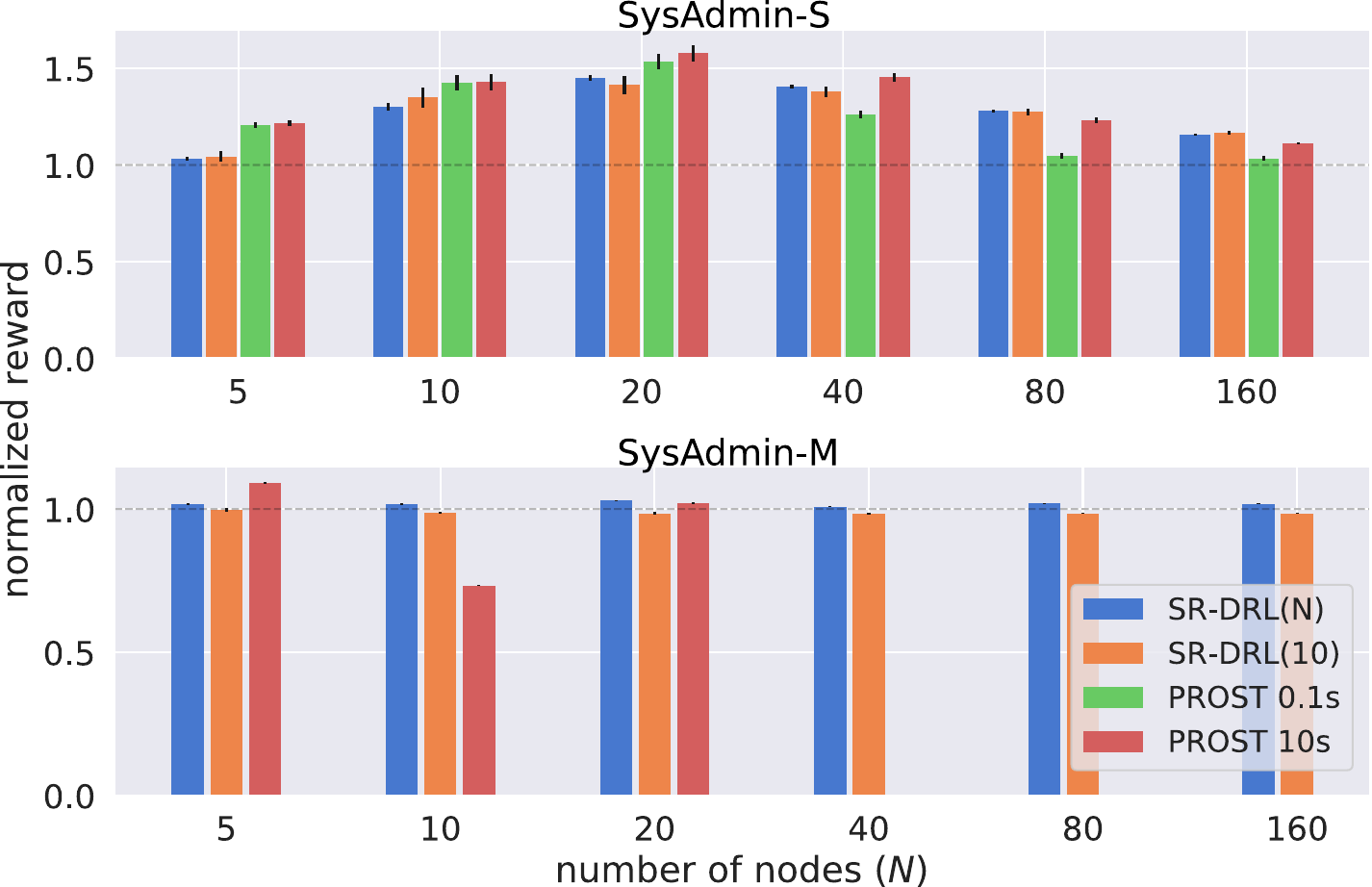} 

  \caption{The figure shows normalized results in SysAdmin-S (a single node can be reset) or SysAdmin-M (multiple nodes can be reset at once). SR-DRL($N$) is trained for the specific $N$, \textbf{SR-DRL(10) is a model trained with $\mathbf{N=10}$ and evaluated in the particular setting}. PROST is a probabilistic planner with either 0.1s or 10s time limit per step. In SysAdmin-M, PROST cannot cannot determine an action in 0.1s, hence we evaluate it only with 10s. Note that SR-DRL takes only about 1ms to decide in both -S and -M variants. Rewards were normalized using baseline algorithms (i.e., to reset a random offline node in SysAdmin-S or all offline nodes in SysAdmin-M at each step).}
  \label{fig:sysadmin}
\end{figure}

\subsubsection{Primary results} \label{sec:sysadmin_results}
For each $N \in \{5, 10, 20, 40, 80, 160\}$, we trained eight SR-DRL agents for 50 epochs (each epoch being 25 600 environment steps). To see how the model generalizes to scenarios with a different number of objects, we picked an agent trained with $N=10$ and evaluated it in different problem variations.
In each setting, we also measured the performance of PROST, an MCTS-based probabilistic planner \citep{keller2012prost,keller2013trial}, with a time allowance of 0.1 or 10 seconds per step. The planner is evaluated with the preset of International Probabilistic Planning Competition 2014. We designed two baseline algorithms that we use to normalize the results: In SysAdmin-S, a random offline node is selected at every step (or none, if there is not any). In SysAdmin-M, all offline nodes are selected for a reset at each step.
On our reference machine, the training time of the SR-DRL algorithm is $15/20/30/50/90/150$ minutes for $N=5/10/20/40/80/160$, respectively. During testing, the model needs about 1 ms per step.

For both SysAdmin-S/-M, all models are evaluated in 100 problem instances with a 100 step limit. The mean results with a 95\% confidence interval are reported in Figure~\ref{fig:sysadmin}. The first observation is that in all instances and both SysAdmin-S/-M, the agent trained with $N=10$ and evaluated for different $N$ performs almost the same as an agent specifically trained in the respective setting. This result indicates that the policy learned for $N=10$ is directly applicable even for $N=160$.

When compared to PROST, the SR-DRL algorithm performs similarly in the SysAdmin-S variation. It performs slightly worse for $N \in \{5, 10, 20\}$. For $N=80$ and $160$, SR-DRL performs slightly better than PROST with 10s per step, while the gap widens when compared to PROST with a 0.1s step limit. We emphasize that our algorithm needs only 1ms per step to decide, after it is trained. That is, for $N \geq 80$ and 100 step limit, PROST-10s needs over 16 minutes to solve the problem, while SR-DRL needs only 0.1 seconds and achieves better performance. This quickly amortizes the training cost, especially since we have shown that a trained model generalizes almost perfectly to different settings.

In SysAdmin-M, the baseline algorithm is very strong, and only PROST-10s with $N=5$ significantly exceeds its performance. Note that with $N=10$, PROST-10s reaches only 73\% of the baseline's performance. However strange, this result was confirmed with several repeated experiments. In this setting, PROST needs to enumerate all action combinations, resulting in $O(2^N)$ space complexity, which becomes infeasible for $N > 20$. Also, even for lower $N$, PROST cannot run with less than 5 seconds per step. Comparatively, the SR-DRL algorithm works well even for $N=160$. 

\subsubsection{Additional experiments}
In Figure~\ref{fig:sysadmin-deps}, we analyzed the behavior of an agent trained in SysAdmin-M to see whether it simply implements the baseline algorithm (i.e., to always reset all offline nodes). We found that the algorithm learns to selectively restart even the online nodes if their failure chance is high (based on the state of their dependencies). However, it seems that the advantage over the baseline algorithm is not significant.

\section{Related work} \label{sec:relatedwork}
\citet{dvzeroski2001relational} introduced RRL in 2001, and approached it with inductive logic programming. As a representative problem, they focused on a more restrictive BlockWorld variant, with only three specific goals: stacking into a single tower, unstacking everything to the ground, or moving a specific box $a$ on top of $b$. For each of these goals, a specialized policy was created, and the authors also reported some degree of generalization to a different number of blocks (3 to 10). However, the used specific goals are trivial compared to the setting we use (any block configuration as a goal). Also, due to a different evaluation procedure, the exact comparison is impossible without carefully reimplementing and evaluating their method.

\citet{payani2020incorporating} replace the hard logic of \citet{dvzeroski2001relational} with a differentiable inductive logic programming. In their approach, the logic predicates are fuzzy, and the parameters are learned with gradient descent. BlockWorld environment is also used for experiments, with an input represented as an image. However, the scope is very limited to only 4 and 5 blocks, without goal generalization (the goal is to always stack into a single tower), and the authors do not report any generalization.

\citet{li2019towards} studies a 3-dimensional instantiation of the BlockWorld problem with a robotic hand and physics simulation. Interestingly, the features of the blocks (their position and color) and the hand are encoded as a graph and processed by a graph neural network (GNN) \citep{zhou2020graph}, making it invariant to the number of objects. Still, all interaction with the world is done by controlling the robotic hand and its elementary actions (relative change of position and grasping controls). Moreover, the blocks are not symbolic, but are identified by their features (color).

Oracle-SAGE \citep{chester2022oracle} investigates how to incorporate planning into our framework. It builds upon the method presented in our paper and explores the idea outlined in the Sokoban experiment -- combining the SR-DRL framework with a planner. It uses our GNN-based models to suggest several macro goals for a planner, uses it to find a sequence of actions to achieve this macro goal and recalculates the next steps. The combined technique performs well in domains that require reasoning over long time or spatial distances. In our work, we present the fundamental principles of the SR-DRL framework without which the Oracle-SAGE would not be able to work.

\citet{hamrick2018relational} study the stability of a tower of blocks in a physical simulation, with some blocks glued together. The blocks and their physical features are encoded as nodes in a graph, and the actions are performed on the graph's edges, which connect two adjoining blocks. This approach generalizes well to different combinations and numbers of blocks. Similarly, \citet{bapst2019structured} focus on the task of creating block structures under physical simulation. They follow a similar approach to ours; their architecture is based on GNN and allows object-centric actions. However, these works focus on their specific domain and do not provide a general framework. Comparatively, we describe a domain-independent framework that works with heterogeneous relations, and multi-parameter actions, possibly with set parameters.

\citet{zambaldi2019deep} and \citet{santoro2017simple} provide specialized neural network architectures with relational inductive biases that internally segment a visual input into objects and process them relationally. Compared to our work, these architectures cannot process symbolic input.

In planning, Relational Dynamic Influence Diagram Language (RDDL) \citep{sanner2010relational} is often used to describe the mechanics of a relational domain. \citet{garg2020symbolic} introduced SymNet, a method that automatically extracts objects, interactions, and action templates from any RDDL. The interacting objects are joined to tuples and represented as a node in a graph; the interactions are represented as edges. A GNN is used to create nodes' embeddings. Action templates, each represented as a single non-linear function, are applied over the object tuples to create a probability distribution. Finally, actions and their parameters (the object tuples) are selected, and the model is updated with a policy gradient method \citep{li2018deep}. There are several drawbacks to the SymNet algorithm. Although the actions share their representation, the final step is to apply the softmax function over all grounded actions. For a problem with $n$ objects and an action with $p$ parameters, this results in $O(n^p)$ time and space complexity. For an action with one set parameter, it is $O(2^n)$. In both cases, the complexity restricts Symnet to problems where actions have only a few parameters. On the other hand, the complexity of SR-DRL is $O(pkn)$, where $k$ is the maximal degree of the graph, which is usually much smaller than $n$. In our case, an action with one set parameter can be computed in $O(n)$, as all nodes are treated independently. The SymNet method is applicable only when the RDDL domain definition is available because it uses the defined transition dynamics to create the graph. In Deep RL, it is common that the transition dynamics are unknown, and only a simulator is available. For example, imagine a visual control domain (e.g., controlling a robotic hand) with automatic object detection \citep[e.g.,][]{redmon2018yolov3}, defined actions manipulating these objects (possibly with a low-level planner in place), and unknown dynamics.

\citet{adjodahsymbolic} studies a control problem of fixed-size maze navigation. The relations are represented as exhaustive binary combinations of all places in the grid and a shared projection is applied to all of them. The output is concatenated and processed with a standard MLP-based Q-learning algorithm. No message passing is involved, allowing the model to reason only with the limited binary relations. Moreover, the learned model is restricted to its specific grid size. 

\citet{groshev2018learning} trains a deep neural network to imitate and generalize behavior generated by a planner. Their neural network architecture is based on image / graph convolutions where the last layer is not fully connected, but it is a fixed-size window centered around the player. This allows generalization to different problem sizes. They apply their method in Sokoban and the traveling salesperson problem.

In visual domains where a relational description is not available, work of \citet{garnelo2016towards} or \citet{zelinka2019building} could be used for automatic object discovery.

The following papers focus on the automatic generation of features from domain descriptions or problem examples to learn generalized policies that transfer to related problems with a different number of objects. \citet{karia2022relational} uses a method from description logic \citep{baader2017introduction} to generate features and learn a generalized Q-function. \citet{ng2021firstorder} also learns a generalized Q-function using a mix of ground and first-order approximations in Relational MDPs \citep{van2005survey}.

\citet{toyer2018action,toyer2020asnets} introduce ASNet, a general neural network architecture that is constructed according to a probabilistic PDDL domain definition and is reusable to all problem instances. This architecture includes interleaved action and proposition layers, which are connected according to the actions' preconditions and effects. Their models output a general reactive policy and are trained through supervised learning based on optimal trajectories determined by an external planner. The authors also tried to train their models with RL, but dismissed this direction due to its inefficiency. Compared to our work, the architecture of our models is semantical (i.e., the graph reflects the objects and their relations) and we use Deep RL for training, which can be applied in domains without known transition dynamics. Moreover, ASNets use a single output for each grounded action, which leads to an exponential number of actions wrt. the number of parameters. We work around this issue with our policy decomposition. 
In a similar work to ASNet, \citet{shen2020learning} takes this a step further and develops a GNN-based method that learns a domain-independent heuristic. 

The work \citep{rivlin2020generalized} introduces a method to tackle domains described in the PDDL language \citep{aeronautiques1998pddl}. They too use GNNs and Deep RL to learn a generalized policy; additionally, they combine the approach with planning. Unlike our work, their approach does not work with multi-parameter actions nor set parameters. \citep{hazra2023deep} uses Deep RL to learn a set of general logic rules for a particular domain. \citep{frances2021learning} learns rule-based policies using combinatorial optimization.

\section{Discussion} \label{sec:discussion}
\subsection{Source of generalization}
The presented experiments demonstrated impressive generalization to problems with a different number of objects and actions. We hypothesize that the source of this generalization lies in the biases induced by the model's architecture -- the GNN and actions operating on the object level. With these, the network can learn general transformations that are transferable between objects. Moreover, as demonstrated in SysAdmin, the close neighborhood of any node may be enough to decide the optimal action for this node. The policy decomposition also helps, because learning several conditional probabilities should be easier than learning a single compound probability. For example, it may be easier to decide what to do next \emph{if} a particular node is selected (i.e., learning $P(a \mid b)$ and $P(b)$ independently), as opposed to learning the compound probability at once (i.e., $P(a, b)$). Moreover, the learned transformation resulting in $P(a \mid b)$ is general and can be immediately applied to all $a \in \mathcal V$ nodes, whereas the transferability of $P(a, b)$ is not clear.

The global node facilitates information transfer over long distances, since it aggregates and then spreads information from all nodes at once. This should help in domains that require non-local, long-distance reasoning. The study of this topic was included in \citep{chester2022oracle}, who found that this approach is limited and may not work in some domains.

\subsection{Architectural choices}
Here we discuss the decisions made regarding the model architecture and training, and their possible modifications.

Our method is designed for policy gradient algorithms, and any algorithm from this family can substitute the used A2C. However, value-based algorithms (e.g., DQN \citep{mnih2015human}) cannot be used, because they do not output probabilistic policy that can be decomposed in the way we described. 

The number of message passes in the GNN has to be tuned for a particular problem. While in BlockWorld, the method requires five (three passes in the GNN phase, and additional two for selecting the second parameter), in Sokoban, it behaves best with ten message passes (see the experiment in Fig.~\ref{fig:sokoban-perf}). This is mainly because Sokoban requires careful planning and the information has to spread over longer distances. 

We tried to keep the model simplistic, keeping the different blocks of the neural network model as single linear layers with LeakyReLU non-linearities when applicable. However, it is possible to use multiple non-linear layers as a replacement for different $\phi$ functions. The possible result could be improved performance with a longer training. 

The used aggregation function in eq.~\eqref{eq:aggregation} is $\max$, but it can be replaced with $\mathrm{mean}$, $\mathrm{sum}$, or, e.g., a concatenation of both $\max$ and $\mathrm{mean}$. In our experiments, $\max$ worked well, but the user may want to try other options.

We used a simple form of attention in the global node to aggregate the information from the whole graph. However, it can attend only for limited information at a time, and hence multi-head attention \citep{vaswani2017attention} could lead to better results in some domains.

\subsection{Domain modeling principles}
The prevalent description of problems is through matrices and elementary actions. However, our world is naturally created of objects and their relations and our framework takes vantage of this fact. Also, low-level planning is a solved problem in a lot of domains (e.g., how to move a robotic hand to a specific position), hence we can focus only on the macro level.

Hence, to use our method to its full potential, we believe that the process of domain modeling should start with identifying its objects, their features and their relations. Let us borrow the cooking example from the introduction. The objects can consist of various ingredients (e.g., a tomato or a piece of meat), tools (a knife, a spoon) and other objects (a cupboard, a frying pan). The relations can code their positions (e.g., the tomato is in the cupboard, the knife lies on a table and the meat is in the pan).

Second, the user has to define the object-centric actions (e.g., open a cupboard, pick the meat or cut the tomato with a knife). 

Finally, the goal can be encoded in a few ways. If it is static, it can be implicitly encoded with the reward function. In other cases, it has to be a part of the input state. In some problems, it can be encoded in the initial global context. When a specific configuration of the world is desired and the model should generalize over different goals, it can be encoded as a separate set of objects and their relations (marked as \emph{goal} relations). This goal graph is then joined with the original input graph (as we do in BlockWorld). In our cooking example, the goal can be a specific configuration in which the meat is cooked and it lies with the tomato on a plate. After the goal is achieved, the agent receives a reward.

The results of our experiments show that the agent generally learns better in smaller problems, but generalizes to larger problems. Hence, it is advisable to take advantage of the fact that the agents can process a variable number of objects and to start small and teach it increasingly complex concepts with curriculum learning \citep{bengio2009curriculum}.

This object representation is human-friendly and can be easily encoded in our framework. Compared to trying to learn this environment through visual input and elementary actions (e.g., moving a hand to a position), it is compact, easier to learn and a lot of the knowledge is transferable between objects, hence the model should generalize better.

\section{Conclusion and future work} \label{sec:conclusion}
We presented a generic framework based on deep reinforcement learning, graph neural networks and autoregressive policy decomposition for solving relational domains. The method operates with a symbolic representation of objects, their relations, and actions manipulating them. We described a generic way to implement multi-parameter actions with mutually dependent parameters, and optionally set parameters that select an arbitrary subset of objects. The action selection operates in linear time and space, w.r.t. the number of objects. One of the great advantages of the framework is that a trained model is not fixed to a specific problem size and can be immediately applied to problems of different sizes.

We demonstrated the framework in three distinct domains, and in all, it showed impressive zero-shot generalization to different problem variations and sizes. In BlockWorld, the model trained solely with five objects solves 78\% of problems with 20 objects, even though the state space grows exponentially with the number of objects. In Sokoban, we show that the method can be joined with a low-level planner and control the environment on its macro-level. When trained solely on 10×10 problems with four boxes, it solves 89\% of 15×15 problems with five boxes. In SysAdmin, once trained model transfers almost perfectly to any other problem size. Moreover, we demonstrated the framework's capability to select multiple objects at once with a single action. Comparatively, a widely used PROST planner cannot work in this setting with a reasonable number of objects due to its exponential complexity.

Future work may explore how to leverage temporal consistency of the graph representations with recurrent neural networks, try to apply curriculum learning \citep{bengio2009curriculum}, fine-tuning to larger domains, or apply the framework to particular problems. For example, when applied to information retrieval, the graph can represent the currently gathered knowledge with actions upon these objects. Similarly, in automated penetration testing, the network objects can be represented as nodes, their network connectivity as relations and actions can define various exploits and scans.

\backmatter
\bmhead{Supplementary information} The complete code for our algorithm in all domains described in this article is publicly available at \url{https://github.com/jaromiru/sr-drl}.

\bmhead{Acknowledgments}
This research was supported by The Czech Science Foundation (grants no. 22-32620S and 22-26655S) and by the OP VVV funded project CZ.02.1.01/0.0/0.0/16\_019/0000765 ``Research Center for Informatics''. This research partially used GPUs donated by the NVIDIA Corporation. Some computational resources were supplied by the project “e-Infrastruktura CZ” (e-INFRA LM2018140) provided within the program Projects of Large Research, Development and Innovations Infrastructures.

\bibliography{citations}

\end{document}